%% file: camera_ready.tex
\documentclass[10pt,twocolumn,letterpaper]{article}

\usepackage{cvpr}              % To produce the CAMERA-READY version
\usepackage{graphicx}
\usepackage{amsmath}
\usepackage{amssymb}
\usepackage{booktabs}
\usepackage{multirow}
\usepackage{color}
\usepackage{lipsum}

\usepackage[accsupp]{axessibility}  % Improves PDF readability for those with disabilities.
\usepackage{algorithm}
\usepackage{algorithmic}
\input{tables.tex}

\input{figures.tex}
\usepackage[pagebackref,breaklinks,colorlinks]{hyperref}

\usepackage[capitalize]{cleveref}
\crefname{section}{Sec.}{Secs.}
\Crefname{section}{Section}{Sections}
\Crefname{table}{Table}{Tables}
\crefname{table}{Tab.}{Tabs.}

\begin{document}

\title{Tencent AVS: A Holistic Ads Video Dataset for Multi-modal Scene Segmentation}
\vspace{-0.3CM}
\author{
      Jie Jiang\thanks{Equal contribution.}, 
      Zhimin Li$^{*}$, 
      Jiangfeng Xiong, 
      Rongwei Quan, 
      Qinglin Lu\thanks{Corresponding author.}, 
      Wei Liu\textsuperscript{\dag}
    \\
    Tencent Data Platform, Shenzhen 518057, China\\
    % \vspace{-0.5cm}
    {\tt\footnotesize  \{zeus, jefxiong, rongweiquan, qinglinlu\}@tencent.com, zhiminli.cn@outlook.com, wl2223@columbia.edu}
}
% }
\maketitle

\begin{abstract}
Temporal video segmentation and classification have been advanced greatly by public benchmarks in recent years. However, such research still mainly focuses on human actions, failing to describe videos in a holistic view. In addition, previous research tends to pay much attention to visual information yet ignores the multi-modal nature of videos. To fill this gap, we construct the Tencent `Ads Video Segmentation'~(TAVS) dataset in the ads domain to escalate multi-modal video analysis to a new level. TAVS describes videos from three independent perspectives as `presentation form', `place', and `style', and contains rich multi-modal information such as video, audio, and text. TAVS is organized hierarchically in semantic aspects for comprehensive temporal video segmentation with three levels of categories for multi-label classification, e.g., `place' - `working place' - `office'. Therefore, TAVS is distinguished from previous temporal segmentation datasets due to its multi-modal information, holistic view of categories, and hierarchical granularities. It includes 12,000 videos, 82 classes, 33,900 segments, 121,100 shots, and 168,500 labels. Accompanied with TAVS, we also present a strong multi-modal video segmentation baseline coupled with multi-label class prediction. Extensive experiments are conducted to evaluate our proposed method as well as existing representative methods to reveal key challenges of our dataset TAVS.
\end{abstract}

%-------------------------------------------------------------------------
%%%%%%%%%%%%%%头图%%%%%%%%%%%%%%

\vspace{-0.2cm}
%%%%%%%%% BODY TEXT
\section{Introduction}
\label{sec:intro}

With the remarkable progress in action recognition approaches~\cite{DBLP:conf/nips/SimonyanZ14,DBLP:conf/eccv/WangXW0LTG16,DBLP:conf/iccv/TranBFTP15,DBLP:conf/eccv/XieSHTM18,DBLP:conf/cvpr/WangL0G18,DBLP:conf/iccv/Feichtenhofer0M19,DBLP:conf/cvpr/Feichtenhofer20,DBLP:journals/corr/abs-2103-15691,DBLP:conf/icml/BertasiusWT21} driven by large-scale benchmarks~\cite{DBLP:conf/cvpr/CarreiraZ17,DBLP:conf/cvpr/HeilbronEGN15,DBLP:conf/cvpr/ShaoZDL20a} in recent years, the {\it de facto} trend of video understanding is to understand human actions by capturing appearance and motion cues. However, few researchers will question whether a single perspective as human action is enough to understand videos given their diverse semantic aspects and multi-modal nature. An important reason to limit comprehensive understanding of videos is the lack of proper benchmarks that incorporate a representative domain of videos with rich semantics beyond actions and build a comprehensive hierarchy of categories.

\figSCENE
\figModalities

Recently, some initial attempts have been made in this direction such as Holistic Video Understanding~(HVU)~\cite{DBLP:conf/eccv/DibaFSPGSG20} and MovieNet~\cite{DBLP:conf/eccv/HuangXRWL20} to overcome the above limitations. HVU collects more annotations from six semantic aspects, e.g., `attribute' and `concept', from existing action recognition datasets~\cite{DBLP:journals/corr/abs-1212-0402,DBLP:conf/cvpr/CarreiraZ17,DBLP:conf/cvpr/GuSRVPLVTRSSM18,DBLP:conf/iccv/KuehneJGPS11,DBLP:conf/iccv/Zhao0TY19}.  It tries to enrich the diversity of video recognition by assigning multiple labels to each video, but it lacks the important temporal analysis process and multi-modal information because it inherits the characteristics of previous datasets. MovieNet further overcomes more limitations of temporal analysis, multi-modal information and categories from multiple aspects by collecting long-form movies. Yet, it still has coarse-level defined segments and insufficient aspects of categories: 1) Its temporal analysis is based on shots detected by traditional methods~\cite{shot-detect} and no accurate temporal annotation is provided, making the temporal segmentation problem a sequential binary classification problem. Thus, the ambiguity of temporal analysis makes it less practical for real-world applications. 2) MovieNet only adds `place' classes into categories, which are even less important than actions for video understanding.

Ideally, we can say that we `understand' a video only if we correctly decompose a video into semantics-consistent parts, recognize semantics of each part, and recover the content of the whole story with another explainable form, e.g., a set of labels. To explore a good representative domain for such high-level intelligence, we believe that ads video, as its properties of rich plot developments and multi-modal nature are just like a `short movie', is a proper arena for video understanding algorithms. In this work, we present the Tencent `Ads Video Segmentation' (TAVS) dataset. In addition to previous datasets which focus on {\em recognition} tasks, e.g., actions or places, we try to incorporate more high-level and abstract classes, e.g., classes from the perspective of video making, to make a step towards {\em cognition} tasks and examine algorithms' ability of learning abstract concepts. For example, we have 7 classes of relationships, e.g., lovers or friends, 6 classes of themes, e.g., family or teaching, and 4 classes of production forms, e.g., animation or cross-cutting. They will be correctly classified only if the algorithms really learn the plot developments across shots among a long-term content in videos. To accommodate the multi-modal nature of videos, e.g., videos, audios, and texts (Fig.~\ref{fig:first}(a)), we also include some classes for modalities other than vision, e.g., dubbing (Fig.~\ref{fig:first}(b)). As a result, a three-level class hierarchy from three perspectives `presentation form', `style', and `place' is built to facilitate the research of video understanding. 

Based on the granularity of our categories, we also define the concept of scene as super-shot in the semantic level with the corresponding granularity in temporal segments. A scene commonly contains one or more consecutive shots and conveys high-level semantics, e.g., the first scene in Fig.~\ref{fig:scene} consists of 4 shots, shifting between the couples in the street and the customer service. Yet, all the 4 shots are about making a phone call, so they are grouped into a scene. By detecting scene boundaries, an effective temporal structure for understanding ads videos is obtained. To make the temporal boundary unambiguous, we also annotate accurate scene boundaries and make our dataset more effective for downstream applications, such as video summarization or editing. Based on the well-defined temporal segments and categories, we believe that our proposed new setting of scene segmentation task, which requires algorithms to temporally segment videos into scenes accurately and predict scene-level multi-label classification results, is a good representative for multi-modal video understanding.

Motivated by the need of the scene segmentation task with multi-modal information, we propose a novel end-to-end `Multi-modal Scene Segmentation Network' (MM-SSN) for simultaneous temporal segmentation and multi-label classification. MM-SSN incorporates multiple prediction heads as dense scene boundary classification, scene boundary refinement, and multi-label classification. It overcomes limitations of previous methods in temporal action segmentation/detection or scene segmentation. The multi-modal feature fusion, temporal modeling, and prediction heads in MM-SSN are integrated to be an end-to-end effective framework for the scene segmentation task.

In summary, our main {\bf contributions} are three-fold: 1) We propose a new setting of scene segmentation task for understanding realistically difficult ads videos with multi-modal information. 2) We build a new benchmark dubbed TAVS. It provides high-quality, temporally accurate, and multi-label annotations with a comprehensive class hierarchy from three perspectives. 3) To reveal key challenges of TAVS, we propose an end-to-end multi-modal video segmentation model and choose proper metrics to evaluate the performances of our proposed baseline and prior representative methods. We conduct extensive ablation studies and would like to facilitate future research on multi-modal video understanding.

\section{Related Works}
\label{sec:related}
\subsection{Video Representation Learning}
\noindent{\bf Single-modal Video Classification.} Video classification is usually investigated in the form of action recognition with single modality. Early efforts including KTH~\cite{DBLP:conf/icpr/SchuldtLC04}, Weizmann~\cite{DBLP:conf/iccv/BlankGSIB05}, UCF-101~\cite{DBLP:journals/corr/abs-1212-0402}, and HMDB~\cite{DBLP:conf/iccv/KuehneJGPS11} commonly have a limited number of categories about human actions. To perform large-scale video representation learning and also serve as a pre-training for downstream tasks, datasets with automatically generated noisy YouTube tags~\cite{DBLP:journals/corr/Abu-El-HaijaKLN16,DBLP:conf/cvpr/KarpathyTSLSF14} or accurate human annotations~\cite{DBLP:conf/cvpr/CarreiraZ17} are constructed for action recognition. Recently, fine-grained action categories with a hierarchy are proposed to reduce the scene bias from backgrounds for action recognition~\cite{DBLP:conf/iccv/GoyalKMMWKHFYMH17,DBLP:conf/cvpr/ShaoZDL20a}. However, their categories (even those with a class hierarchy) are still limited to human actions and lack the holistic view of videos, such as the style and presentation form of videos. They also ignore the multi-modal information and have no temporal annotations, so ours dataset is different from them.

\noindent{\bf Multi-label Video Classification.} Previous datasets are mainly resulted from untrimmed videos with concurrent or temporally independent actions, such as Multi-MiT~\cite{DBLP:journals/corr/abs-1911-00232}, Charades~\cite{DBLP:conf/eccv/SigurdssonVWFLG16}, ActivityNet~\cite{DBLP:conf/cvpr/HeilbronEGN15}, MPII-Cooking~\cite{DBLP:conf/cvpr/RohrbachAAS12}, and MultiTHUMOS~\cite{DBLP:journals/ijcv/YeungRJAMF18}. The noisy annotations generated from YouTube tags~\cite{DBLP:journals/corr/Abu-El-HaijaKLN16,DBLP:conf/cvpr/KarpathyTSLSF14} also result in some videos with multiple classes. Our dataset has accurate multi-label annotations from three perspectives to achieve a holistic view to understand ads videos, therefore different from the previous ones. The most similar multi-label dataset to ours in the video classification area is HVU~\cite{DBLP:conf/eccv/DibaFSPGSG20} with six different tasks for classification. However, it does not provide multi-modal contents and temporal annotations for video segments, so we address a different need from it.

\noindent{\bf Multi-modal Video Representation Learning.} Early efforts for supervised multi-modal video classification mainly explore the usage of rich web data with image, video, audio, and text and construct their own datasets~\cite{DBLP:conf/mm/LinH02,DBLP:journals/www/TianTPCS19,DBLP:journals/sigpro/LiuFZ16,DBLP:journals/tmm/JiangWTLXC18}, yet there are few public benchmarks for evaluation. Some other attempts exploit audio or text information from public video classification benchmarks to enrich video representations~\cite{DBLP:journals/corr/abs-2001-08740,DBLP:journals/corr/WangKRMSYSCETZZ17}. Recently, large-scale unsupervised multi-modal video datasets without categories are proposed to learn better representations for downstream tasks~\cite{DBLP:conf/iccv/SunMV0S19,DBLP:conf/iccv/MiechZATLS19,DBLP:journals/corr/abs-2101-10803}, which show superior performance over supervised counterparts on many downstream tasks, e.g., action recognition~\cite{DBLP:conf/cvpr/MiechASLSZ20,DBLP:conf/nips/AlayracRSARFSDZ20}, action segmentation~\cite{DBLP:conf/cvpr/ZhuY20a,DBLP:conf/cvpr/MiechASLSZ20}, and video retrieval~\cite{DBLP:conf/cvpr/ZhuY20a,DBLP:conf/cvpr/MiechASLSZ20,DBLP:conf/nips/AlayracRSARFSDZ20,DBLP:conf/eccv/Gabeur0AS20}. Our dataset provides human annotated class labels and temporal boundaries for the video segmentation task, so it is different from the previous datasets.

\subsection{Temporal Video Segmentation}
\label{sec:segmentation}
\noindent{\bf Temporal Action Segmentation Datasets.} These datasets commonly contain instructional videos~\cite{DBLP:conf/cvpr/KuehneAS14,DBLP:conf/huc/SteinM13,DBLP:conf/cvpr/TangDRZZZL019} with human performing fine-grained actions and require algorithms to predict an action label for each frame with only visual information. Our dataset differs from them by providing rich multi-modal information and requiring multi-label predictions for each video segment. 

\noindent{\bf Scene Segmentation Datasets. } Its major difference from action segmentation is the shot-based boundary detection setting, which means that videos are split into shots in the process of dataset construction, and the shot-based boundary detection task is actually transformed into a sequential binary classification problem, e.g., whether each shot boundary is a scene boundary. Typical datasets are about short videos~\cite{DBLP:journals/ijsc/RotmanPA17} or documentaries~\cite{DBLP:conf/mm/BaraldiGC15}, and recently scaled up to long-form movies~\cite{DBLP:conf/cvpr/RaoXXXHZL20}. They commonly contain multi-modal information, such as place, actor, action, and audio, but they lack the important category hierarchy for classifying each segment in order to comprehensively understand videos and also lack the accurate annotations of segment boundaries. Therefore, this task is different from ours and is not applicable to our task, either.

\noindent{\bf Temporal Modeling in Video Segmentation}. Scene segmentation methods~\cite{DBLP:conf/cvpr/RaoXXXHZL20} often use an LSTM~\cite{DBLP:journals/neco/HochreiterS97} with shot-level multi-modal inputs to predict the binary classification result for each shot boundary as scene boundary. This type of temporal modeling cannot be applied to our dataset with frame-level temporal annotations. Action segmentation methods typically use temporal modeling networks such as encoder-decoder~\cite{DBLP:conf/cvpr/LeaFVRH17,DBLP:conf/cvpr/LeiT18} or dilated convolutions~\cite{DBLP:conf/cvpr/FarhaG19} upon off-the-shelf feature extractors~\cite{DBLP:conf/cvpr/CarreiraZ17} to predict a single action for each frame. Recent approaches~\cite{DBLP:conf/wacv/IshikawaKAK21,DBLP:conf/eccv/WangGWLW20} also incorporate action boundary detection methods to handle over-segmentation errors. Our proposed end-to-end baseline is built upon MS-TCN++~\cite{DBLP:journals/corr/abs-2006-09220} in action segmentation by adding the multi-modal fusion module to handle multiple modalities as input and the extra prediction heads to account for multi-label classification, scene boundary classification, and scene boundary refinement, respectively.

\section{The TAVS Benchmark}
\label{sec:benchmark}
In this section, we will introduce our Tencent Ads Video Segmentation (TAVS) benchmark. To make it solid, we collect large-scale ads videos from real industry data and annotate scene boundaries and class labels with careful quality control. We also examine existing metrics and choose proper ones to evaluate the performance on our benchmark.

\subsection{Ads Video Segmentation Dataset}
\label{sec:dataset}
Based on our defined `scene' in Sec.~\ref{sec:intro}, we build our large-scale TAVS dataset with accurate annotations for scene boundaries and scene-level multi-label categories.

\noindent{\textbf{Data Collection}}. Our ads videos are collected from real industry data, where their diversity is manually controlled according to the domains, such as education, e-business, video game, finance, and Internet service. Different from previous cooking/instructional actions in the action segmentation task or movies in the scene segmentation task, we believe that ads videos from various industrial domains will provide more diverse scenarios to test the performances of algorithms, e.g., people moving or talking in the education domain and artificial scenes in video games.

\figTaxonomy

\noindent{\textbf{Taxonomy Generation.}} By carefully inspecting the patterns of collected ads videos, we believe that ads videos are as `short' movies since they were also filmed from the script and edited by editors. Thus, their key characteristics could be summarized from advertiser-uploaded scripts. By exploiting the statistics of words' frequencies in scripts, we summarize hundreds of key words by deleting trivial words from high-frequency words. Then, we use K-means~\cite{macqueen1967some} to cluster them and get several centers like `place', `presentation', `character', and `style' to be the class hierarchy v1.0. Based on domain knowledge of making ads videos, we further add more classes in the production process, e.g., shooting forms such as `interview' or `multi-person sitcom' for filming original videos and generation forms such as `crossing cutting' or `slide show' for editing ads videos. For abstract concepts in `style', we also add classes of emotions to let them complete. As a result, we build a three-layer hierarchy from three perspectives: `presentation form', `style' and `place'. After removing the classes with a very small number of instances (i.e., $\leq$ 100) when we finish the annotation process, our TAVS dataset has 82 high-frequency classes, as shown in Fig.~\ref{fig:class}. The numbers of categories with the presentation, style, and place are 25, 34, and 23, respectively. Fig.~\ref{fig:dist} shows the label distribution of presentation, style, and place with different colors, which is a typical long-tailed distribution. Please refer to Appendix B.2 for the complete list of our categories.

%For example, in presentation forms, there are covering grid layout, dubbing, or padding; in styles, ad videos contains various emotions or relationships; in places, there are home, office, or school. 
\figPerClassStat

\Structuringtable
\noindent{\textbf{Annotation Process and Quality Control}}. Dissimilar to the existing action localization or segmentation task whose main challenge in the annotation process is the ambiguity of action boundaries, human-edited ads videos have many shot changes as clear demarcations. Therefore, the main challenge of our dataset is the annotations of our multi-label fine-grained categories. Specifically, a scene has about 6 classes on average, so annotators easily miss some classes instead of annotating a wrong label. Based on these concerns, we write a detailed handbook for annotators, which contains an elaborated detailed description and a sample video for each class, and ask them to pay attention to the easily-missed classes. Our handbook also describes when scene changes and in which situation the scene crosses multiple shots. In our annotation process, we first let 7 annotation companies try to annotate 100 videos and select top 3 companies to annotate our entire dataset according to their annotation accuracies. During annotation, we randomly sample about 1/10 videos for each batch and check the quality of annotations, especially the missing classes. We approve the batch if the accuracy of annotation is satisfying (e.g., $\geq$ 90\%), and otherwise let the annotators revise their annotations of the whole batch. Each video's annotations will be revised for about 3 times on average. We also adopt heuristic rules, e.g., mutually exclusive classes should not appear in the same video, to examine the quality. As a result, we believe that the quality of our dataset is carefully controlled. Finally, as scene boundaries annotated by humans are hard to reach per-frame precision, we use a shot detection method TransNet v2~\cite{DBLP:journals/corr/abs-2008-04838} to refine shifting of scene boundaries as post-processing, i.e., we use the nearest shot boundary to substitute the original scene boundary annotation if their distance is small (i.e., $\leq$ 0.1s). Please refer to Appendix B.1 for visualization of our annotation tool and our guidebook for annotators.
%\figDuration

\tableCLS

\noindent{\textbf{Access to The TAVS Dataset.}} Our dataset will be made publicly available once our paper is accepted. We will release annotations of the train/val set and all 12,000 videos, and maintain an online server for evaluating the performance on the test set.

\subsection{Dataset Statistics}
\label{sec:stat}
Ads videos in the TAVS dataset come from real industry data. It contains 12,000 Ads videos (totally 142.1 hours) and is divided into 5,000 for the training set, 2,000 for the validation set, and 5,000 for the test set. Most of video durations in TAVS are between 25 and 60 seconds, with average length as 45.5 sec. In Fig.~\ref{fig:dist}, per-class data size shows that TAVS has a typical long-tailed distribution, indicating the realistic difficulty of our dataset. As shown in Table~\ref{tab:compare_structuring}, previous multi-modal scene segmentation datasets commonly have no class annotations (e.g., MovieScenes~\cite{DBLP:conf/cvpr/RaoXXXHZL20}), while TAVS has. Previous single-modal action segmentation datasets commonly have no shot changes in videos and only require single-label predictions, and most of them are smaller than TAVS (e.g., 50Salads~\cite{DBLP:conf/huc/SteinM13} and Breakfast~\cite{DBLP:conf/cvpr/KuehneAS14}). Furthermore, our dataset covers a wider range of real-world scenarios (e.g., city street or office) instead of a very small domain in action segmentation (e.g., cooking). Thus, our class hierarchy is more diverse and richer than previous ones. In Table~\ref{tab:compare_cls}, we compare TAVS with other multi-label video recognition datasets. The scale of most previous datasets is much smaller than ours, while HVU~\cite{DBLP:conf/eccv/DibaFSPGSG20} has no temporal annotations and no multi-modal information. MovieNet~\cite{DBLP:conf/eccv/HuangXRWL20} has more categories than ours, but it merely contains `action' and `place' classes while our TAVS has many high-level abstract classes. The scale of MovieNet is also smaller than ours.

\subsection{Metrics}
\label{sec:metrics}
Due to the properties of our scene segmentation task, e.g., multi-label in the temporal segmentation setting, many metrics adopted by existing tasks cannot directly be used in our task. Here we first briefly review some related metrics and then introduce the metrics we adopted. 

\noindent{\textbf{Metrics of previous tasks.}} The most related action segmentation task~\cite{DBLP:conf/cvpr/LeaFVRH17,DBLP:conf/cvpr/FarhaG19,DBLP:conf/eccv/WangGWLW20} adopts per-frame accuracy, edit-distance based score, and F1-score with different tIoUs. However, the multi-label annotation of our task makes these metrics hard to compute and unsuitable to use. Previous scene segmentation methods~\cite{DBLP:conf/cvpr/RaoXXXHZL20} without scene-level class annotations use mAP of scene boundaries as their metric. However, they commonly first use a shot detection method~\cite{shot-detect} and regard the scene segmentation task as a binary classification task to determine whether each shot boundary is a scene boundary or not. Therefore, it is also unsuitable for our accurate scene boundary annotations with 0.04s (per-frame annotation in 25fps videos) as minimum units. 

\noindent{\textbf{Metrics of our TAVS task.}} To measure the performance of our TAVS task, we first adopt the mAP@tIoU metric similar to temporal action detection task (e.g., ActivityNet~\cite{DBLP:conf/cvpr/HeilbronEGN15}) to evaluate the performance of multi-label classification and IoU-based localization. We also adopt the F1@distance metric similar to the GEBD challenge~\cite{DBLP:journals/corr/abs-2101-10511} to evaluate the accuracy of scene boundary localization solely. Specifically, for mAP@tIoU, we require the submitted scene segments to have no overlaps to adapt the detection metrics~\cite{DBLP:conf/cvpr/HeilbronEGN15} to our scene segmentation task. We take an average of mAPs from IoU 0.5 to 0.95 with stride 0.05 as our final mAP evaluation. For F1@time, we use absolute distances instead of relative ones in~\cite{DBLP:journals/corr/abs-2101-10511} to avoid the influence from long segments to their neighboring short segments. Formally, the average mAP we use is
\vspace{-0.8em}
\begin{equation}
\small
Avg\_mAP=\frac{1}{10} \sum_{t I o U=0.5}^{0.95} \frac{1}{N} \sum_{i=1}^{N} A P_{i}^{t I o U}, 
\end{equation}
where $N$ is the total number of categories, and $AP$ is the Average Precision commonly used in detection tasks. To measure the accuracy of scene boundary localization, we use F1-score from 0.1s to 0.5s, named as Avg\_F1. For computing F1-score in each temporal distance, we match each predicted boundary to the nearest ground-truth scene boundary and regard it as a True Positive (TP) if their distance is less than $t$, otherwise it is a False Positive (FP). The ground-truth scene boundary will be deleted once it is positively matched (i.e., $\leq t$), and the ground-truths which do not have any positive matches to be False Negatives (FN). Therefore, the precision is TP/(TP+FP) and recall is TP/(TP+FN). The final Avg\_F1 is computed by 
\vspace{-0.5em}
\begin{equation}
\small
Avg\_F1 = \frac{1}{5}\sum_{t=0.1}^{0.5}\frac{2 \cdot Precision_t \cdot Recall_t}{Precision_t + Recall_t}.
\end{equation}

\subsection{Dataset Characteristics}
As discussed above, our proposed new setting of scene segmentation task and the TAVS benchmark have several distinguishing characteristics compared against existing tasks or datasets.

\noindent{{\bf Difficulty}}. TAVS is difficult in several aspects comparing to previous related datasets: 1) Multi-label classification from a holistic hierarchy, while previous ones are limited to only have multiple overlapped actions~\cite{DBLP:journals/ijcv/YeungRJAMF18} or no temporal annotations~\cite{DBLP:conf/eccv/DibaFSPGSG20}; 2) rich multi-modal information as video frame, audio, text from ASR and OCR, while previous ones commonly contain single visual modality. The realistic difficulty of TAVS will be a good arena for algorithms to understand multi-modal videos as well as our real world.

\noindent{{\bf High Quality}}. Our videos are collected from real industry data and are with high-resolutions. With our careful quality control, the class labels and temporal segments are accurately annotated. In addition, compared to MovieNet~\cite{DBLP:conf/eccv/HuangXRWL20} which only releases three frames per shot, the richer content of original videos will be released in our TAVS dataset.

\noindent{{\bf Real-world Applications}}. Due to the realistically difficult ads videos in our TAVS dataset, it has a great potential for applications. For example, 1) Creative video editing: by understanding the detailed content of ads videos from different perspectives as in our TAVS dataset, a single ads video can be automatically split into segments and re-organized to have various lengths, which will save the efforts of video editors; 2) recommendation: the fine-grained categories of each video segment rather than the coarse video-level categories can better serve ads recommendation. Please refer to Appendix C.1 for the discussion of our future work.

\figBaseline

\section{Multi-modal Scene Segmentation Network}
Based on the discussion on existing scene segmentation or action segmentation methods in videos (Sec.~\ref{sec:segmentation}), we find that no existing methods are suitable for our proposed multi-modal ads video segmentation task. Thus, we propose our baseline model `Multi-modal Scene Segmentation Network' (MM-SSN) for multi-modal and multi-label video segmentation, as shown in Fig.~\ref{fig:baseline}. Our model exploits three modalities in our TAVS dataset and performs multi-modal fusion operations to integrate all the information. Then, a temporal modeling network with a similar architecture to MS-TCN++~\cite{DBLP:journals/corr/abs-2006-09220} is used to capture long-term dependencies. At last, each location with integrated long-term information is used to perform three predictions: multi-label classification, scene segmentation, and scene boundary regression.

\noindent{{\bf Feature Extractors}}. We use three off-the-shelf feature extractors for modalities in our MM-SSN: images with Swin Transformer~\cite{DBLP:journals/corr/abs-2103-14030}, audio with VGGish~\cite{DBLP:conf/icassp/HersheyCEGJMPPS17}, and text with BERT~\cite{DBLP:conf/naacl/DevlinCLT19}. Due to the fact that either our categories or the contents of our videos have little connection with motions, the performance of the pretrained video feature backbone S3D~\cite{DBLP:conf/eccv/XieSHTM18} on the HowTo100M dataset~\cite{DBLP:conf/iccv/MiechZATLS19} is even worse than the strong image backbone Swin Transformer trained on the large-scale ImageNet dataset~\cite{DBLP:conf/cvpr/DengDSLL009}. For better computational efficiency, we choose the image feature in 2fps instead of video feature in our MM-SSN. Texts in ads videos are extracted by ASR~\cite{DBLP:conf/icassp/ZhangLYD18} with noisy timestamps. We approximate these texts to 2fps images according to their timestamps and use zero-padding in those moments without people talking.

\noindent{{\bf Multi-modal Fusion}}. As the importance of each modality is not equal, we adopt a two-stage fusion strategy for three modalities: 1) for well-aligned frames and audio features with accurate timestamps, we first concatenate frame features $T \times D_1$ and audio features $T \times D_2$ as $T \times (D_1 + D_2)$, and then perform a channel attention operation similar to SE~\cite{DBLP:conf/cvpr/HuSS18} to let salient features among two modalities automatically be chosen; 2) the salient features with positional embeddings (PE) will be fused with text features (also with PE) by cross-attention~\cite{DBLP:conf/nips/VaswaniSPUJGKP17}. Then, the fused feature is concatenated with the salient feature in Stage 1 to form a final multi-modal feature.

\noindent{{\bf Temporal Modeling}}. To capture the long-term dependencies of videos, we adopt MS-TCN++~\cite{DBLP:journals/corr/abs-2006-09220} to greatly enlarge the receptive field of our model by dilated convolutions. Specifically, MS-TCN++ has $N$ stages and each stage has $L$ dilated convolutional layers with dilation $r=2^l$, where $l$ is the number of the current layer, and $N$ and $L$ are hyper-parameters. For the first stage, MS-TCN++ has extra $L$ layers with dilation being $r=2^{L-l}$ to simultaneously capture neighborhood and remote information.

\noindent{{\bf Prediction Heads and Loss Functions}}. Previous temporal action segmentation methods such as MS-TCN++~\cite{DBLP:journals/corr/abs-2006-09220} commonly utilize a cross-entropy loss for frame-wise classification and a smoothing loss for regularization, respectively. Thus, the boundary of each predicted segment is decided according to the transition of classification results, where no explicit boundary localization is performed. To overcome the challenge of multi-label classification in our dataset, we disregard the common practice in action segmentation tasks and develop three heads for predicting segment boundaries and their multi-label categories. Before performing segment boundary localization and regression, we adopt a temporal difference module in adjacent frames to make features more salient for segment boundary localization. A sequential binary cross-entropy is used to classify whether each sampled location is a scene boundary. The accurate ground-truth is approximated into a 0-1 sequence according to the sampling rate (e.g., 2 fps). We also use a boundary regression head to tackle the tiny temporal errors by predicting the offsets from the nearest sampled location to the ground-truth timestamp by a smooth-$L_1$ loss. For classification head, we adopt the ASL loss~\cite{DBLP:journals/corr/abs-2009-14119} on the output of the temporal modeling module for multi-label classification, which has promising performance on long-tailed distributions and is thus suitable for our dataset. We use it with multiple temporal scales: we first use all sampled locations, and then we temporally perform max-pooling of predicted classification confidence scores inside each scene (determined by ground-truth) or inside the whole video and use ASL for scene-level or video-level classification to further boost the performance. Please refer to Appendix A.1 for the implementation details of our MM-SSN.

\section{Experiments}
We comprehensively investigate representative methods on the related tasks and present ablation studies of our proposed MM-SSN on our TAVS benchmark. Results are reported in the test set by selecting the best model on the validation set.
\subsection{Scene Segmentation Results}
In this section, we compare our proposed MM-SSN with representative single-modal temporal action detection or segmentation methods, and multi-modal scene segmentation methods on our TAVS dataset. Please refer to Appendix A.2 for implementation details.

\figPerClassAP
\noindent{{\bf Temporal Action Detection}: BMN~\cite{DBLP:conf/iccv/LinLLDW19} + NeXtVLAD~\cite{DBLP:conf/eccv/Lin0018}}. BMN is a representative and effective temporal action detection method to generate action proposals, thus we choose it to predict segment boundaries by replacing the `start' and `end' classes with only one `boundary' class. Then, the segments will be classified into multi-label categories by a representative video classification method NeXtVLAD. 

\noindent{{\bf Temporal Action Segmentation}: MS-TCN++~\cite{DBLP:journals/corr/abs-2006-09220}} is a representative temporal action segmentation method to perform frame-wise single-label action classification. We simply replace the cross-entropy loss with a multi-label classification loss, i.e., Binary Cross Entropy~(BCE) loss, and use a threshold of $0.5$ to generate segments. We use a simple post-processing strategy to merge labels for the same segment.

\noindent{{\bf Scene Segmentation}: LGSS~\cite{DBLP:conf/cvpr/RaoXXXHZL20} + NeXtVLAD~\cite{DBLP:conf/eccv/Lin0018}}. Since there are shot changes in our TAVS dataset, we can also adopt a traditional scene segmentation method to predict scenes and then perform multi-label classification for each scene. LGSS is a representative method for scene segmentation by using multi-modal LSTM to aggregate shots (resulted from shot detection~\cite{shot-detect}) with a sequential binary classification loss.

\tableSOTA
\tableModality
% \begin{figure*}[t]
%     \setlength{\abovecaptionskip}{-0.0cm}
%         \setlength{\belowcaptionskip}{-0.0cm}
%     \centering
%     \includegraphics[width=1.0\textwidth]{figures/ap.pdf}
%     \caption{Per-class scene-level AP analysis by comparing our multi-modal baseline with its single-modal variants on the TAVS benchmark.}
%     \label{fig:ap}
%     % \vspace{-2em}
% \end{figure*}
\tableFusion

\noindent{{\bf Comparison of Results}}: We adopt the same feature encoders for fair comparisons with our MM-SSN: 1) single modality: Swin Transformer for BMN, NeXtVLAD, and MS-TCN++; 2) multiple modalities: Swin Transformer, VGGish, and BERT for LGSS. In Tab.~\ref{tab:sota}, we compare the performance of MM-SSN to these methods. Even with an independent multi-label classification network, BMN still generates much worse proposals than our MM-SSN, as shown in F1 scores. We believe the reason is that its 2D-proposal map limits the number of its fixed-length sampling, so the localization errors tend to be larger than 0.5s. MS-TCN++ also achieves a poor performance on our dataset because its segment boundary localization is performed by the transition of classification results, which is not suitable for the studied multi-label segmentation problem. Due to the inaccurate shot detector, LGSS has much worse performance than MM-SSN and also has more complex multi-stage architectures.

\subsection{Ablation Study}
\noindent{{\bf How important is each modality?}} As a multi-modal dataset, modalities are important for the final performance. As shown in Tab.~\ref{tab:modality}, video frame is the most important modality as expected, while all three modalities could further improve the overall performance of mAP by 1\%. We believe that it is a considerable improvement based on the strong performance of our baseline architecture with optimized prediction heads and loss functions.
%\tableLoss

% \figPerClassAP}{
% \begin{figure*}[t]

% \end{figure*}
% }

\noindent{{\bf Which category is more challenging?}} As shown in Fig.~\ref{fig:dist}, our dataset has a long-tailed distribution, which makes tail classes even more challenging. In Fig.~\ref{fig:ap}, we perform a per-class scene-level AP analysis (tIoU 0.5-0.95) by comparing MM-SSN with its single-modal variants (only video frames). We observe that the performance of classes is not strictly aligned with class sizes. Tail classes like `studio' and classes closely related to other modalities like `dubbing' tend to benefit more from multi-modal inputs. 
\tableTemporal
\tableBoundary
\tableRate
\noindent{{\bf Multi-modal Fusion Strategies}}. In Tab.~\ref{tab:fusion}, we compare some variants with our two-stage fusion strategy: 1) simple concatenation; 2) SE for all modalities; 3) SE for video and audio features, and then self-attention for fusing the text feature by concatenating it along the temporal dimension; 4) SE and dual-direction cross-attention; 5) SE and single-direction cross-attention. Although their results are close to each other, our fusion strategy still gets a small performance gain, especially in mAP.

\noindent{{\bf Temporal Modeling}}. In Tab.~\ref{tab:temporal}, we compare representative temporal modeling techniques: transformer encoder achieves the best results in F1@0.5, but it makes features less discriminative and therefore the result of Avg\_mAP is much lower. Bi-LSTM achieves the worst result due to its poor ability to handle long sequences in videos.

\noindent{{\bf Scene Boundary Localization Strategies}}. We design two components to enhance the ability of accurate scene boundary localization. As shown in Tab.~\ref{tab:bd}, the boundary regression head greatly contributes to higher F1-score, while the feature difference module also leads to a notable performance gain in F1@0.1s. In Tab.~\ref{tab:rate}, we also investigate the influence of sampling rate: 5 fps leads to a considerable improvement in F1 yet a little decrese in mAP. For both high mAP and high computational efficiency, we choose 2 fps as default.

\noindent{{\bf Classification Losses}}. Due to the long-tailed distribution of our dataset, directly using a multi-label classification loss, such as the BCE loss, will lead to sub-optimal performance. We compare the recent ASL loss~\cite{DBLP:journals/corr/abs-2009-14119} with BCE: the ASL loss obtains a notable performance gain in mAP as about 1.5\% and almost the same results in F1 score.

\section{Conclusion}
Our goal of this work is to drive the research of video understanding in novel directions by recommending a new scenario of scene segmentation task for study to the computer vision community. Accompanied with a specially collected Tencent Ads Video Segmentation (TAVS) benchmark, we would believe that the proposed task on this TAVS benchmark can lead to expected or unexpected innovations. We reveal the key challenges in TAVS and would like to provide useful guidance for future research. We would also hope that the proposed task and the TAVS dataset will catalyze deeper research in video-related areas, leading to exciting new breakthroughs.

\clearpage
{\small
\bibliographystyle{ieee_fullname}
\bibliography{camera_ready}
}

\clearpage
\end{document}

%% file: tables.tex
\usepackage{pifont}
\newcommand{\tableCLS}{
\begin{table}[t]
\setlength{\abovecaptionskip}{-0.0cm}
        \setlength{\belowcaptionskip}{-0.0cm}
        \caption{Comparison with multi-label video recognition datasets.}
        \centering
    \resizebox{0.48\textwidth}{!}{
        \scalebox{1.0}{
            \begin{tabular}{r|lllll}
            \toprule
            \textbf{Dataset}       & \begin{tabular}[c]{@{}l@{}}\textbf{\#Labels}\end{tabular} & \textbf{\#Videos} & \begin{tabular}[c]{@{}l@{}}\textbf{\#Labels}\\ \textbf{per video}\end{tabular} & \textbf{\#Classes} &  \textbf{Domain}                        \\ \midrule
            Charades~\cite{DBLP:conf/eccv/SigurdssonVWFLG16}   & 67k             & 9.8k     & 6.8                                                        & 157                    & Actions at Home  \\
            MPII-Cooking~\cite{DBLP:conf/cvpr/RohrbachAAS12} & 13k                                                          & 273 & 46                                                         & 78               & Cooking          \\
            MultiTHUMOS~\cite{DBLP:journals/ijcv/YeungRJAMF18} & 38.7k                                                        & 400 & 10.5                                                       & 65                    & Sports           \\
            HVU~\cite{DBLP:conf/eccv/DibaFSPGSG20} & 7,499k                 & 572k & 13.1                                                       & 3142                    & Sports           \\
            MovieNet~\cite{DBLP:conf/eccv/HuangXRWL20} & 64.6k                                    & 1.1k & 58.7                                                      & 170                    & Movies           \\ \midrule
            \textbf{TAVS} & \textbf{168.5k}                                                       & \textbf{12k}    & \textbf{14.1}                                                      & \textbf{82}                & \textbf{Ads}        \\ \bottomrule
            \end{tabular}
        }
    }
    \label{tab:compare_cls}
\end{table}
}

\newcommand{\Structuringtable}{
\begin{table*}[t]
\setlength{\abovecaptionskip}{-0.0cm}
        \setlength{\belowcaptionskip}{-0.0cm}
        \caption{Comparison with video segmentation datasets. Action segmentation datasets annotate per-frame action classes, but they commonly have no multi-modal information. Scene segmentation datasets only annotate scene boundaries but no scene-level classes. TAVS has rich multi-modal inputs and holistic scene-level categories. $\dagger$ means using ASR during annotation, yet no multi-modal inputs. }
        \centering
    \resizebox{1.0\textwidth}{!}{
    \scalebox{1.0}{
        \begin{tabular}{r|c|lllllll}
        \toprule
        \textbf{Dataset} & \textbf{Anno. Type}                   & \textbf{Duration} & \textbf{\#Video} & \textbf{\#Shot} & \textbf{\#Segment} & \textbf{\#Class}  & \textbf{Multi-modal}& \textbf{Domain} \\ \hline
        MPII-Cooking~\cite{DBLP:conf/cvpr/RohrbachAAS12}             & \multirow{9}{*}{\begin{tabular}[c]{@{}c@{}}Action\\ segmentation\end{tabular}}         & 9.8h              & 44               & -               & 5.6k               & 65                & \ding{55} & cooking actions                         \\
        YouCook~\cite{DBLP:conf/cvpr/DasXDC13}          &                                 & 2.4h              & 88               & -               & -                  & 10                & \ding{55} & cooking actions                         \\
        50Salads~\cite{DBLP:conf/huc/SteinM13}         &                                 & 5.4h              & 50               & -               & 996                & 17                & + Depth & cooking actions                         \\
        Breakfast~\cite{DBLP:conf/cvpr/KuehneAS14}        &                                 & 77h               & 1.7k             & -               & 8.4k               & 48               & \ding{55} & cooking actions                         \\
        `5 tasks'~\cite{DBLP:conf/cvpr/AlayracBASLL16}        &                                 & 5h                & 150              & -               & -                  & 47                & \ding{55}$\dagger$ & instructional videos                        \\
        Ikea-FA~\cite{DBLP:conf/dicta/ToyerCHG17}          &                                 & 3.8h             & 101              & -               & 1.9k               & 13                & \ding{55} & assembling furniture                       \\
        YouCook2~\cite{DBLP:conf/aaai/ZhouXC18}         &                                 & 176h              & 2k             & -               & 13.8k              &  89              & \ding{55} & cooking actions                         \\
        Epic-Kitchens~\cite{DBLP:journals/corr/abs-1804-02748}    &                                 & 55h               & 432              & -               & 39.6k              & 397               & \ding{55}$\dagger$ & cooking actions                         \\
        COIN~\cite{DBLP:journals/corr/abs-1903-02874}             &                                 & 476.5h            & 11.8k            & -               & 46.3k              &  180             & \ding{55} & instructional videos                        \\ \midrule
        OVSD~\cite{DBLP:journals/ijsc/RotmanPA17}            & \multirow{3}{*}{\begin{tabular}[c]{@{}c@{}}Scene\\ Segmentation\end{tabular}} & 10h               & 21               & 10k           & 3k               & -                & + Audio/ASR & Movie                                     \\
        BBC~\cite{DBLP:conf/mm/BaraldiGC15}              &                                 & 9h                & 11               & 4.9k            & 670                & -                & + ASR & Documentary                                       \\
        MovieScenes~\cite{DBLP:conf/cvpr/RaoXXXHZL20}      &                                 & 297h               & 150              & 270k            & 21.4k              & -                & + Audio & Movie                                      \\ \midrule
        \multirow{2}{*}{\begin{tabular}[c]{@{}c@{}}\textbf{TAVS}\end{tabular}}      & \multirow{2}{*}{\begin{tabular}[c]{@{}c@{}}{\bf Scene}\\ {\bf Segmentation}\end{tabular}}                           & \multirow{2}{*}{\begin{tabular}[c]{@{}c@{}}\textbf{142.1h}\end{tabular}}            & \multirow{2}{*}{\begin{tabular}[c]{@{}c@{}}\textbf{12k}\end{tabular}}              & \multirow{2}{*}{\begin{tabular}[c]{@{}c@{}}\textbf{121.1k}\end{tabular}}            & \multirow{2}{*}{\begin{tabular}[c]{@{}c@{}}\textbf{33.9k}\end{tabular}}              & \multirow{2}{*}{\begin{tabular}[c]{@{}c@{}}\textbf{82}\end{tabular}}               & \multirow{2}{*}{\begin{tabular}[c]{@{}c@{}}\textbf{+ Audio/ASR/OCR}\end{tabular}} & \multirow{2}{*}{\begin{tabular}[c]{@{}c@{}}\textbf{Advertisement}\end{tabular}}                             \\
        &&&&&&&&\\
        \bottomrule
        \end{tabular}
            }
        }
        \label{tab:compare_structuring}
\end{table*}
}

\newcommand{\tableSOTA}{
\begin{table}[t]
\caption{Comparison with previous representative methods on related tasks. $^\dagger$ not originally designed for our task.}
\centering
\setlength{\abovecaptionskip}{0.0cm}
        \setlength{\belowcaptionskip}{-0.4cm}
\resizebox{0.48\textwidth}{!}{
\begin{tabular}{l|ccccc}
\toprule
Methods &Multi-modal &\multicolumn{1}{l}{F1@0.1s}  &  \multicolumn{1}{l}{F1@0.5s}& \multicolumn{1}{l}{Avg\_F1} & \multicolumn{1}{l}{Avg\_mAP} \\ 
\midrule
BMN~\cite{DBLP:conf/iccv/LinLLDW19} + NeXtVLAD~\cite{DBLP:conf/eccv/Lin0018}  &  &   11.81   &    64.51      & 38.78&27.94         \\
MS-TCN++~\cite{DBLP:journals/corr/abs-2006-09220}$^\dagger$        &        &   1.65     &  5.12    &     3.76            &      0.12            \\
MM-SSN (single-modal)  &   & \textbf{42.06}   &   81.61     &   69.40              &     29.06 \\ \midrule
LGSS~\cite{DBLP:conf/cvpr/RaoXXXHZL20} + NeXtVLAD~\cite{DBLP:conf/eccv/Lin0018} & \checkmark&  29.41   &  31.52   &30.78   &13.95           \\ 
Our MM-SSN & \checkmark & 41.95        & \textbf{82.72} &  \textbf{69.69}                   & \textbf{30.99}           \\ 
\bottomrule
\end{tabular}
}
\label{tab:sota}
\end{table} 
}

\newcommand{\tableModality}{
\begin{table}[t]
\setlength{\abovecaptionskip}{-0.0cm}
        \setlength{\belowcaptionskip}{-0.0cm}
\caption{Ablation on modalities used in our baseline.}
\centering

\resizebox{0.48\textwidth}{!}{
\begin{tabular}{l|cccc}
\toprule
Modalities &\multicolumn{1}{l}{F1@0.1s} &  \multicolumn{1}{l}{F1@0.5s}&  \multicolumn{1}{l}{Avg\_F1} & \multicolumn{1}{l}{Avg\_mAP} \\ 
\midrule
Video frame   & \textbf{42.06}   &   81.61     &   69.40              &     29.06                   \\ 
Audio      &  20.03  & 53.78 &   39.57               &     11.30                     \\ 
Text         &  19.44  & 50.12 &      38.53           &     7.04                    \\ 
All (MM-SSN)   & 41.95       & \textbf{82.72} &  \textbf{69.69}                   & \textbf{30.99}      \\ 
\bottomrule
\end{tabular}
}
\label{tab:modality}

\end{table} 
}

\newcommand{\tableFusion}{
\begin{table}[t]
\setlength{\abovecaptionskip}{-0.0cm}
        \setlength{\belowcaptionskip}{-0.0cm}
\caption{Ablation on fusion strategies.}
\centering
\resizebox{0.48\textwidth}{!}{
\begin{tabular}{l|cc}
\toprule
Fusion Strategies &  \multicolumn{1}{l}{Avg\_F1} & \multicolumn{1}{l}{Avg\_mAP} \\ 
\midrule
Concat (3 modalities)        &   69.97    &   29.31                   \\ 
SE (3 modalities)                    &        \textbf{70.95}         &     30.37    \\ 
SE (video + audio) + self-attn (+ text)       &     69.68        &           30.76    \\ 
SE (video + audio) + dual cross-attn (+ text) & 69.16           & 30.34               \\ 
SE (video + audio) + cross-attn (+ text)      &     69.69      &   \textbf{30.99}      \\ 
\bottomrule
\end{tabular}
}

\label{tab:fusion}
\end{table} 
}

\newcommand{\tableTemporal}{
\begin{table}[t]
\setlength{\abovecaptionskip}{-0.0cm}
        \setlength{\belowcaptionskip}{-0.0cm}
\caption{Ablation on temporal modeling methods.}
\centering

\resizebox{0.48\textwidth}{!}{
\begin{tabular}{l|cccc}
\toprule
Temporal Modeling &  \multicolumn{1}{l}{F1@0.1s} &  \multicolumn{1}{l}{F1@0.5s}&  \multicolumn{1}{l}{Avg\_F1}& \multicolumn{1}{l}{Avg\_mAP} \\ 
\midrule
Bi-LSTM              &       36.70     &  81.84    &     66.97       &      27.31       \\ 
Transformer Encoder  &    35.88        &  \textbf{83.04}    &   69.01      &    21.64               \\ 
MS-TCN++ (in MM-SSN)             & \textbf{41.95}        & 82.72 &  \textbf{69.69}                   & \textbf{30.99}            \\
\bottomrule
\end{tabular}
}
\label{tab:temporal}
\end{table} 
}

\newcommand{\tableBoundary}{
\begin{table}[t]
\setlength{\abovecaptionskip}{-0.0cm}
        \setlength{\belowcaptionskip}{-0.0cm}
\centering
\caption{Ablation on the scene boundary localization module.}
\resizebox{0.48\textwidth}{!}{
\begin{tabular}{cc|cccc}
\toprule
reg. head & feat. diff.  &  \multicolumn{1}{l}{F1@0.1s} &  \multicolumn{1}{l}{F1@0.5s}&  \multicolumn{1}{l}{Avg\_F1}& \multicolumn{1}{l}{Avg\_mAP} \\ 
\midrule
                 &       &  22.00     &    80.72      &  55.54   &  28.71        \\
                  & \checkmark& 22.14& 81.74 &56.03 &28.29\\
   \checkmark &           & 39.49     & 82.29 &   69.06                   & 28.47   \\ 

  \checkmark  &    \checkmark  & \textbf{41.95}        & \textbf{82.72} &  \textbf{69.69}   & \textbf{30.99}     \\ 

\bottomrule
\end{tabular}
}
\label{tab:bd}
\end{table} 
}

\newcommand{\tableRate}{
\begin{table}[t]
\setlength{\abovecaptionskip}{-0.0cm}
        \setlength{\belowcaptionskip}{-0.0cm}
\caption{Ablation on the sampling rate of video frames.}   
\centering
\resizebox{0.48\textwidth}{!}{
\begin{tabular}{c|cccc}
\toprule
Sampling rate &  \multicolumn{1}{l}{F1@0.1s} &  \multicolumn{1}{l}{F1@0.5s}&  \multicolumn{1}{l}{Avg\_F1}& \multicolumn{1}{l}{Avg\_mAP} \\ 
\midrule
1 fps                 &      26.11          &   74.98          &     55.01           &       29.15              \\
2 fps                & 41.95        & 82.72 &  69.69   & \textbf{30.99}           \\ 
5 fps                 & \textbf{66.52}        & \textbf{82.80}     &   \textbf{77.85}             &   29.16         \\ 
\bottomrule
\end{tabular}
}
\label{tab:rate}
\vspace{-0.3cm}
\end{table} 
}

%% file: figures.tex
\newcommand{\figModalities}{
\begin{figure}[t]
\setlength{\abovecaptionskip}{-0.03cm}
        \setlength{\belowcaptionskip}{-0.6cm}
\centering
\includegraphics[width=0.47\textwidth]{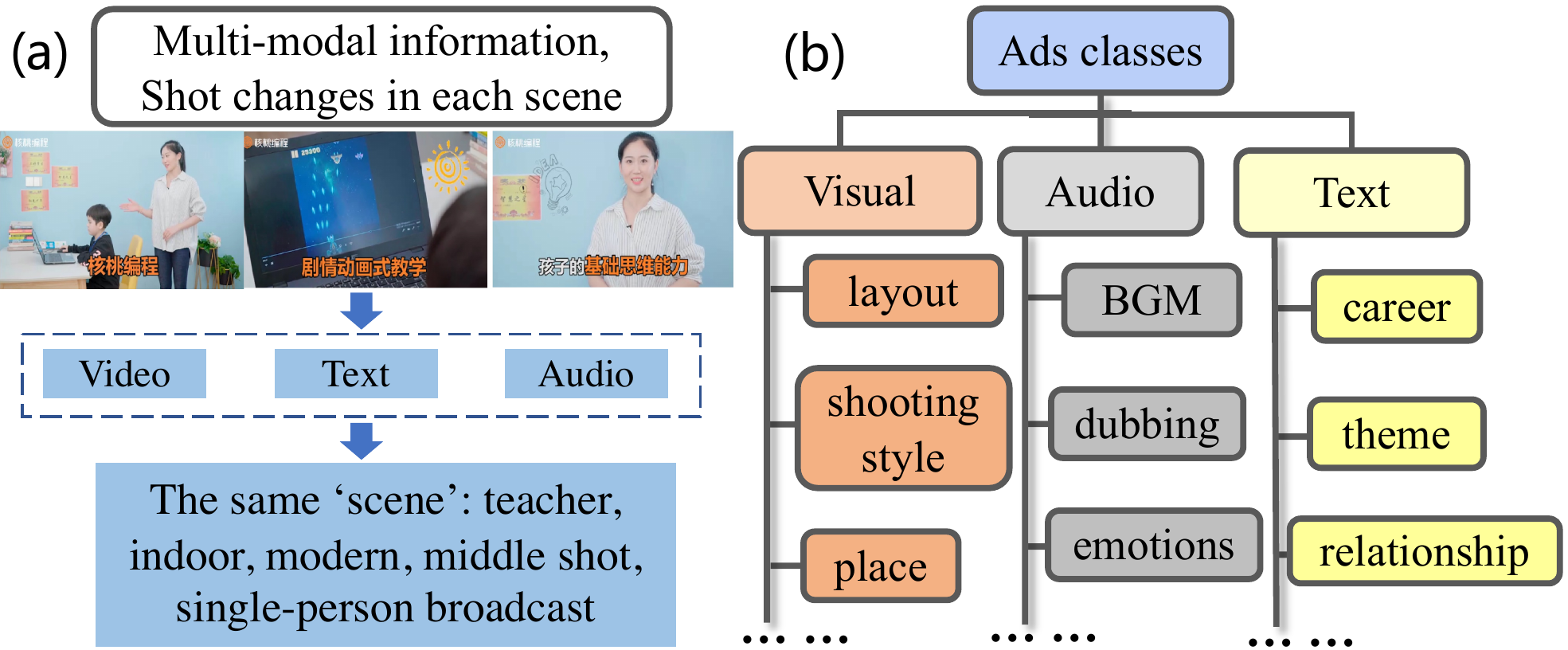}
\vspace{0.5cm}
\caption{Visualization of challenges in the TAVS dataset: (a) multi-modal information, and scene composed of shots with different contents; (b) multi-modal classes.}
\label{fig:first}
\end{figure}
}

\newcommand{\figSCENE}{
\begin{figure*}[t!]
	\begin{center}
		\includegraphics[width=1.0\textwidth]{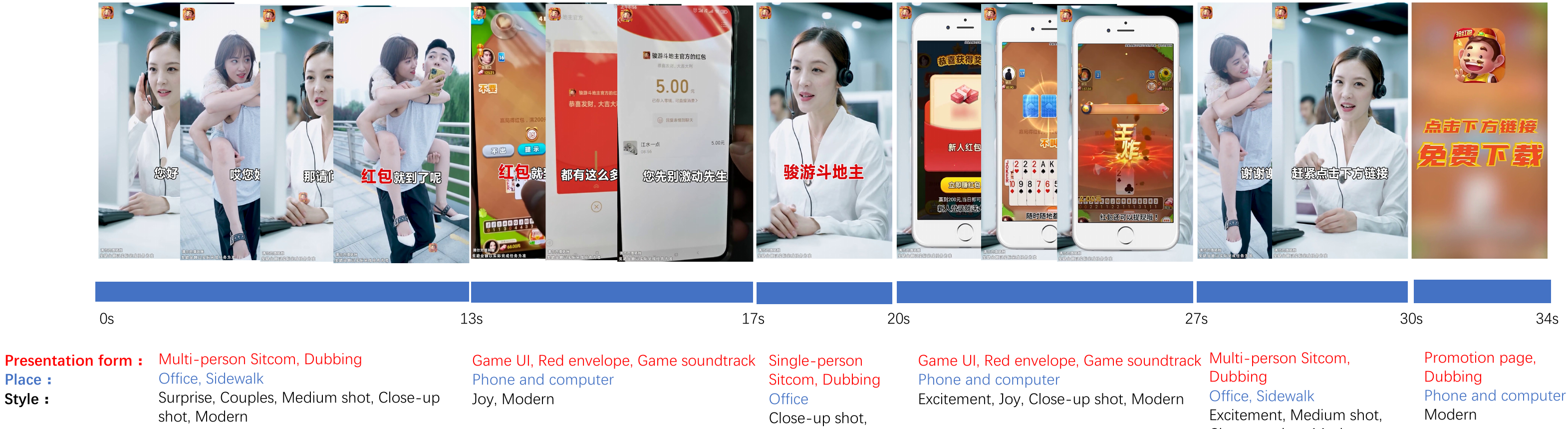}
	\end{center}
	\caption{Visualization of a typical ads video in the TAVS dataset. Our new setting of scene segmentation requires precisely segmenting videos as scenes and predicting scene-level multi-label categories.}
	\label{fig:scene}
\end{figure*}
}

\newcommand{\figPerClassStat}{
\begin{figure*}[t]
\setlength{\abovecaptionskip}{-0.03cm}
        \setlength{\belowcaptionskip}{-0.0cm}
    \centering
    \includegraphics[width=1.0\textwidth]{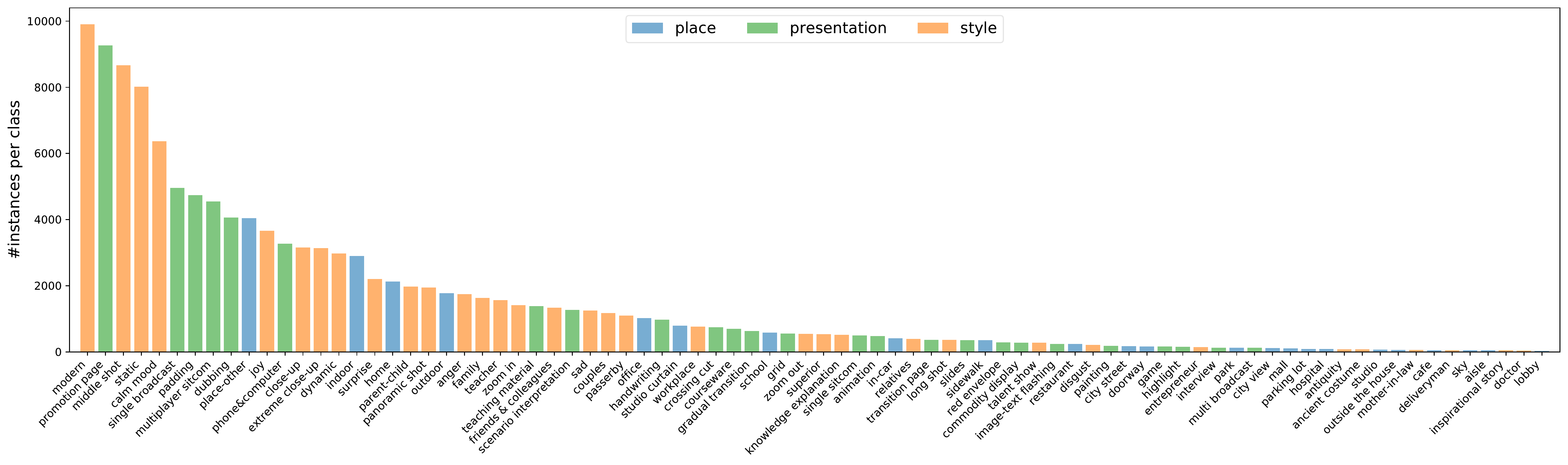}
    \caption{Statistics of per class data size. The three perspectives of our multi-label classification are in different colors.}
    \label{fig:dist}
\end{figure*}
}

\newcommand{\figTaxonomy}{
\begin{figure}[t]
\setlength{\abovecaptionskip}{-0.00cm}
\setlength{\belowcaptionskip}{-0.7cm}
    \centering
    \includegraphics[width=0.45\textwidth]{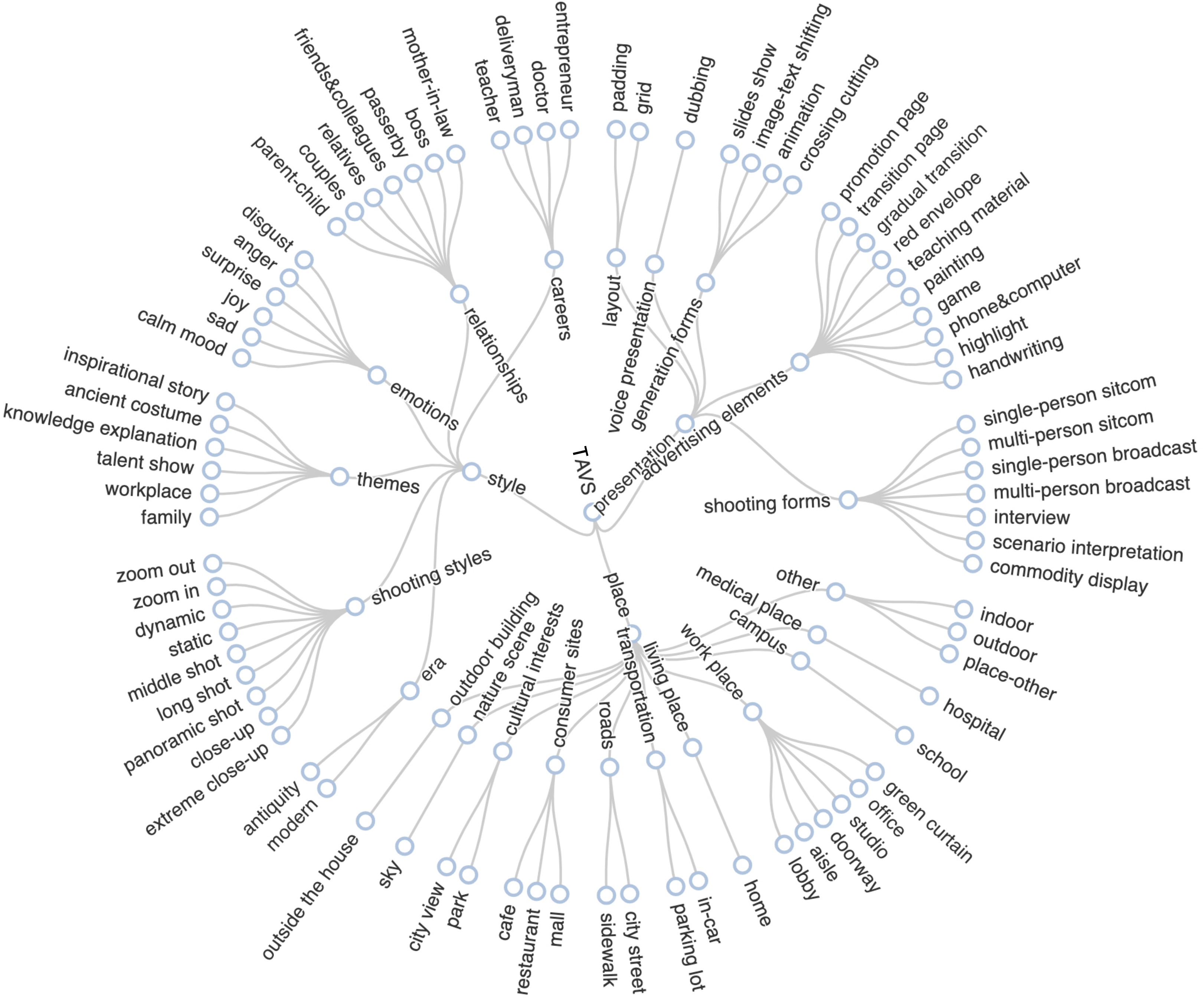}
    \caption{The taxonomy tree of the TAVS dataset.}
    \label{fig:class}
\end{figure}
}

\newcommand{\figBaseline}{
\begin{figure*}[t]
\setlength{\abovecaptionskip}{-0.0cm}
        \setlength{\belowcaptionskip}{-0.2cm}
   \centering
   \includegraphics[width=1\textwidth]{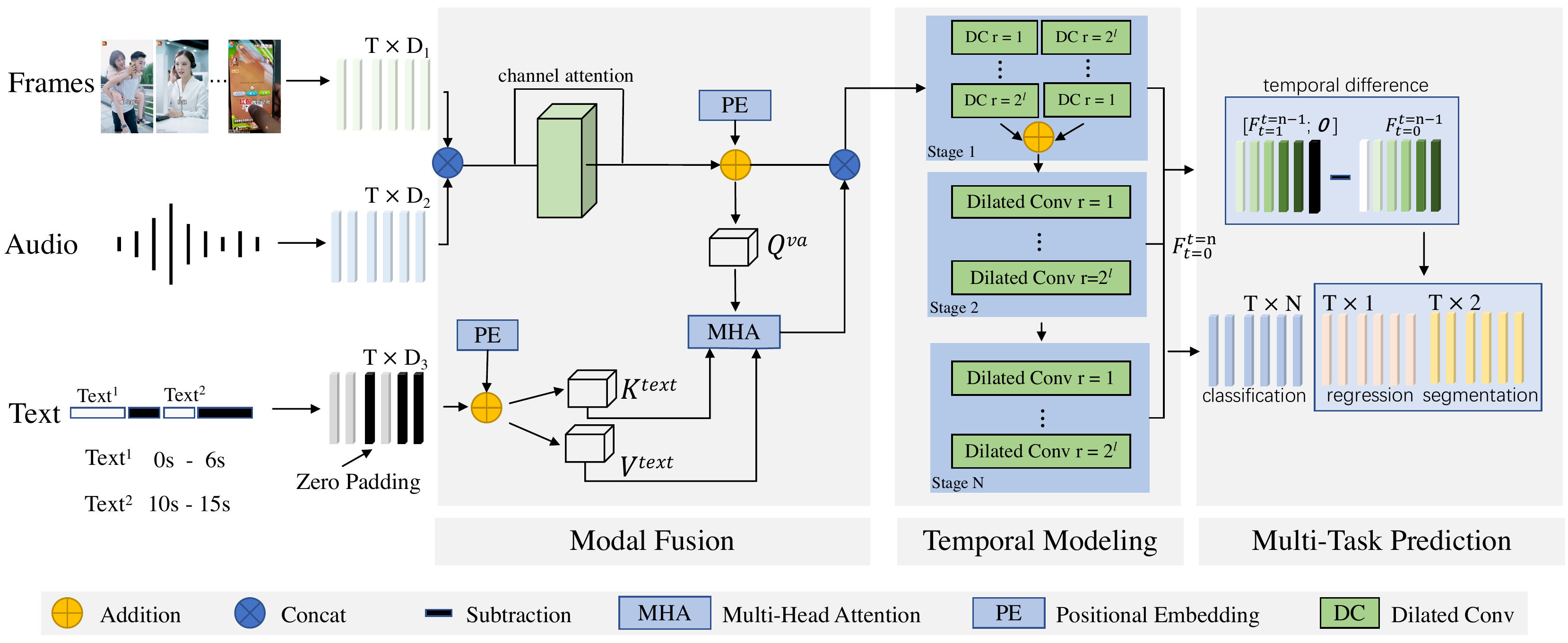}
   \caption{The architecture of our proposed multi-modal baseline.}
\label{fig:baseline}

\end{figure*}
}

\newcommand{\figPerClassAP}{
\begin{figure*}[t]
\setlength{\abovecaptionskip}{-0.0cm}
        \setlength{\belowcaptionskip}{-0.0cm}
    \centering
    \includegraphics[width=1.0\textwidth]{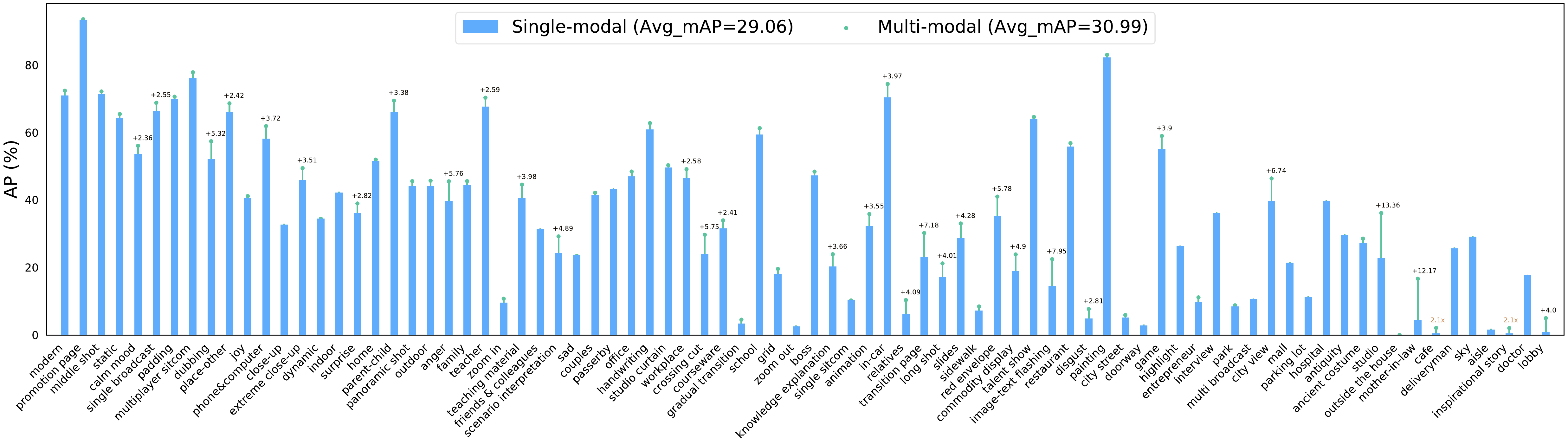}
    \caption{Per-class scene-level AP analysis by comparing our multi-modal baseline with its single-modal variants on the TAVS benchmark.}
    \label{fig:ap}
\end{figure*}
}